\documentclass{article}

\PassOptionsToPackage{numbers, compress}{natbib}

\usepackage[preprint]{neurips_2023}
\usepackage[utf8]{inputenc} 
\usepackage[T1]{fontenc}    
\usepackage{hyperref}       
\usepackage{url}            
\usepackage{booktabs}       
\usepackage{amsfonts}       
\usepackage{nicefrac}       
\usepackage{microtype}      
\usepackage{xcolor}         
\usepackage{multirow}
\usepackage{makecell}
\usepackage{comment}
\usepackage{graphicx}
\usepackage{wrapfig}
\usepackage{caption}
\usepackage{color, colortbl}
\usepackage[titletoc,toc,page]{appendix}
\definecolor{mygray}{gray}{.9}

\title{UniBoost: Unsupervised Unimodal Pre-training for Boosting Zero-shot Vision-Language Tasks}

\author{%
  Yanan Sun$^1$, Zihan Zhong$^2$, Qi Fan$^1$, Chi-Keung Tang$^1$, Yu-Wing Tai \\
  $^1$The Hong Kong University of Science and Technology, 
  $^2$Tsinghua University \\
  \texttt{now.syn@gmail.com},  \texttt{zhongzh22@mails.tsinghua.edu.cn}, 
  \texttt{fanqics@gmail.com}, \\ 
  \texttt{cktang@cs.ust.hk}, \texttt{yuwing@gmail.com} 
}

\begin{document}
\maketitle

\begin{abstract}
Large-scale joint training of multimodal models, e.g., CLIP, have demonstrated great performance in many vision-language tasks. However, image-text pairs for pre-training are restricted to the intersection of images and texts, limiting their ability to cover a large distribution of real-world data, where noise can also be introduced as misaligned pairs during pre-processing. Conversely, unimodal models trained on text or image data alone through unsupervised techniques can achieve broader coverage of diverse real-world data and are not constrained by the requirement of simultaneous presence of image and text. In this paper, we demonstrate that using large-scale unsupervised unimodal models as pre-training can enhance the zero-shot performance of image-text pair models. Our thorough studies validate that models pre-trained as such can learn rich representations of both modalities, improving their ability to understand how images and text relate to each other. Our experiments show that unimodal pre-training outperforms state-of-the-art CLIP-based models by 6.5\% (52.3\% $\rightarrow$ 58.8\%) on PASCAL-5$^i$ and 6.2\% (27.2\% $\rightarrow$ 33.4\%) on COCO-20$^i$ semantic segmentation under zero-shot setting respectively. By learning representations of both modalities, unimodal pre-training offers broader coverage, reduced misalignment errors, and the ability to capture more complex features and patterns in the real-world data resulting in better performance especially for zero-shot vision-language tasks. 
\end{abstract}

\section{Introduction}
\vspace{-0.1in}
Vision-language tasks have attracted much attention, producing excellent results by capitalizing on language-supervised pre-trained models such as CLIP~\cite{CLIP}. Motivated by the strong generalization ability of CLIP, extensive research effort has been made to transfer the knowledge in these language-supervised pre-trained models to downstream tasks, e.g., open-vocabulary detection~\cite{ViLD, OV-DETR} and semantic segmentation~\cite{zsseg, ZegFormer, lseg}. 

However there are two main limitations of such image-text pair models. One is the limiting size of image-text pairs that restricts their ability to cover a large distribution of real-world data. The other limitation is that they require a significant amount of pre-processing, where errors can be introduced in aligning image and text data which hinder their accuracy. Conversely, unimodal models that are trained on text or image data alone can achieve a much broader coverage of the real-world distribution of data, see Figure~\ref{fig:teaser}. This is because these models are not constrained by the requirement that image and text data to be present simultaneously for pair-up.  Unimodal models are often trained on larger datasets with much less pre-processing and no pair-ups, and thus errors caused.  

\begin{wrapfigure}{R}{0.5\textwidth}
\vspace{-0.15in}
  \centering
  \includegraphics[width=0.48\textwidth]{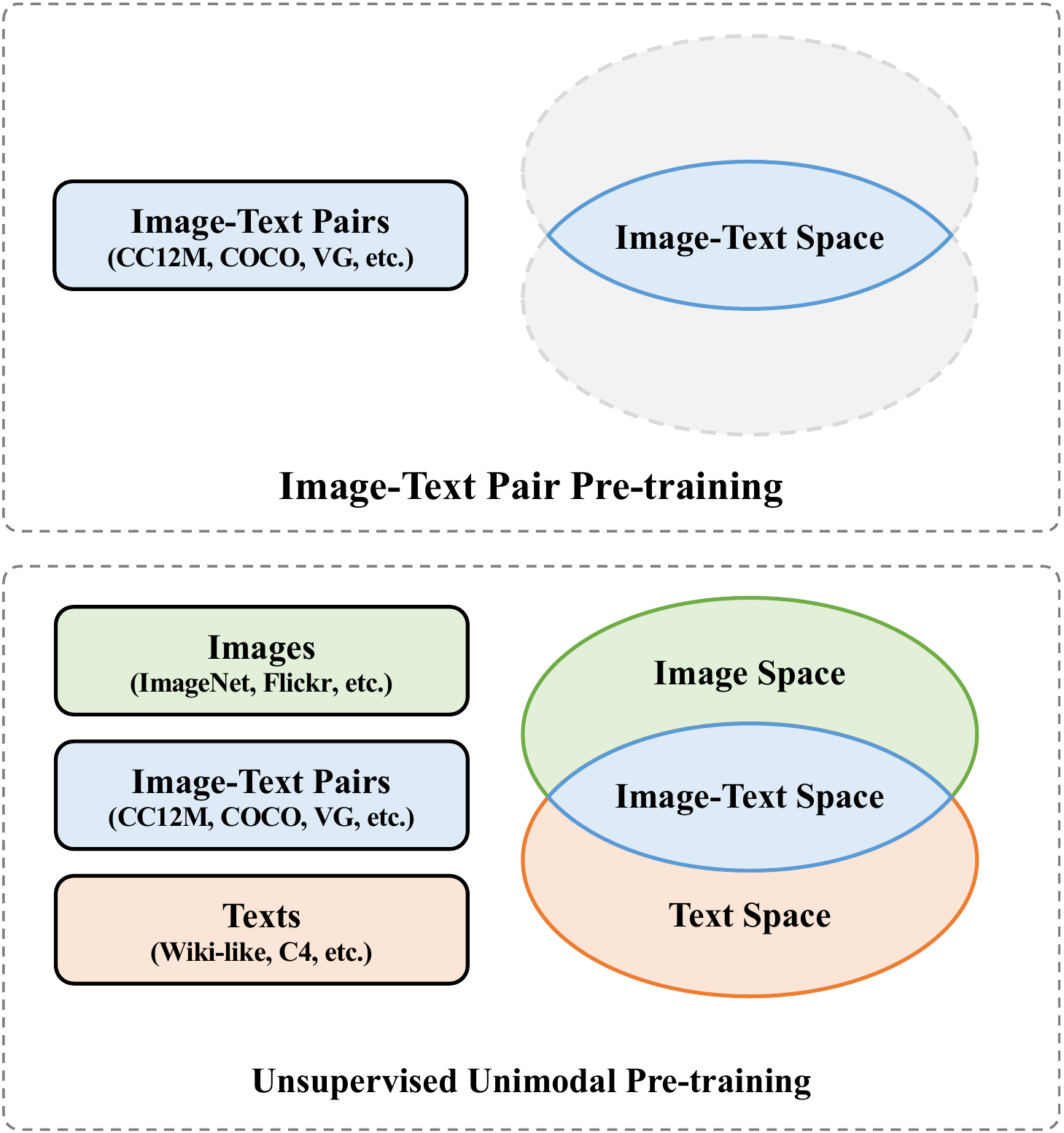}
  \caption{The solution space of conventional vision-language pre-training is restricted to image-text pairs while unsupervised unimodal models can be trained on not only image-text pairs, but also image or text data alone, i.e., a broader range of data distribution. Our UniBoost, a multitask intermediate fine-tuning framework based on unsupervised unimodal supervised models, benefit from the general and robust representations of unsupervised unimodal models and thus boost zero-shot vision-language tasks.}
  \vspace{-0.2in}
\end{wrapfigure}\label{fig:teaser}

In this paper, we find that {\em unsupervised unimodal pre-training} is surprisingly more effective than supervised pre-training or image-text pair supervised (or language-supervised) pre-training under the same experimental setup. Specifically, 
zero-shot segmentation based on two unimodal pre-trained weights, namely, MAE~\cite{MAE} and T5~\cite{T5}, outperforms the models based on CLIP-Image and CLIP-Text by more than 5\% on mIoU under similar model capacity.
Supervised pre-training or language-supervised pre-training learns the representative embeddings based on paired data, e.g., image-text pairs, that is, learns from overlapped image-text data space. In contrast, unsupervised pre-training can learn from data more easily {\em without} any of these requirements. 

Current trend of vision-language tasks adopts pretrain-then-finetune paradigm, where the model is first pre-trained from scratch on large-scale image-text pairs, and then fine-tuned on specific task. During pre-training, the models learn the multimodal representation and align the image-text space on image-text pairs simultaneously. 
In contrast, we propose to learn from unimodal pre-trained models for vision-language tasks. Unimodal pre-trained models can make use of all available image and text data rather than limited to image-text pairs, which are then aligned or fused through in-domain fine-tuning on downstream tasks. During in-domain fine-tuning, the respective image and language spaces are correlated to produce multimodal representation for vision-language tasks.

Despite the prospect of covering broader distribution, our experiments show that it is challenging to train multimodal fusion modules based on unimodal images and text encoders on limited data, which is difficult to be generalized to new tasks and domains. In this paper, we design a multitask intermediate fine-tuning framework based on unsupervised unimodal pre-trained models called {\bf UniBoost} which encompasses the following contributions. First, UniBoost can help leverage multiple related tasks to learn more general and robust representations. The model is forced to learn to extract useful features from multiple sources and integrate them effectively. Second, UniBoost can help to regularize the model and prevent overfitting, as the model is exposed to a wider range of data and tasks during intermediate fine-tuning. Finally, UniBoost helps to improve the efficiency and scalability of the model, as it can be trained on larger and more diverse datasets without requiring additional labeled data. This can be especially beneficial when working with limited labeled data.

\vspace{-0.1in}
\section{Related Work}
\vspace{-0.1in}
\noindent\textbf{Vision-Language Models (VLMs)}  aim to bridge the gap between visual and textual modalities. VLM has been intensively investigated recently on various multimodal tasks, e.g., visual question answering (VQA)~\cite{peng2020mra,hu2019language}, image/video captioning~\cite{pan2004automatic,kulkarni2013babytalk,li2019entangled,zhang2019object}, visual grounding~\cite{mao2016generation,yu2016modeling,liu2017referring}, referring segmentation~\cite{ye2019cross,feng2021encoder,ding2021vision} and text-to-image generation~\cite{reed2016generative,gu2022vector}. Recently, vision tasks such as instance segmentation~\cite{lseg, Fusioner} and object detection~\cite{ViLD, DetPro} introduce text embeddings into the model designs to boost performance or implement open-vocabulary pixel or object classification.
Specifically, VLM pre-training~\cite{CLIP,jia2021scaling,yu2022coca,ofa,flamingo,FLIP,VLMo,beit3} has provided strong and effective foundation models for these multimodal applications. The VLMs are pre-trained to learn rich vision-language correspondences from large-scale informative image-text pairs supervised by certain vision-language objectives~\cite{MoCo,BERT,yang2022unified,MAE,singh2022flava,li2022grounded}. The pre-trained high-capacity VLMs can perform zero-shot predictions on downstream applications even without fine-tuning by matching the feature embeddings of images and texts directly.
However, despite their success,  existing VLMs are trained on image-text pairs of the intersection of image and text data. By contrast, we propose to explore the union of image and text data for VLMs, which is fundamentally and theoretically much larger than the image-text pairs data. Different from ~\cite{VLMo, beit3} utilizing a multi-way transformer to pre-train on image-text pair data and image or text data alone from scratch, we directly transfer the knowledge from well pre-trained unimodal models and learn the multimodal alignment or fusion through intermediate fine-tuning on in-domain tasks, which is more efficient and flexible.

\noindent\textbf{Unsupervised Pre-training (UPT)}
fuels the progress of VLM foundation models by enabling effective usage of massive internet data for model pre-training. It significantly reduces the dependency on data labeling, where the encoder networks are trained with self-supervision by specific pretext tasks.
Early UPT works design special pretext tasks and train the model to predict the corresponding answers, e.g., context prediction~\cite{doersch2015unsupervised}, inpainting~\cite{pathak2016context}, colorization~\cite{zhang2016colorful}, jigsaw~\cite{noroozi2016unsupervised}, visual primitives counting~\cite{noroozi2017representation}, and rotation prediction~\cite{komodakis2018unsupervised}.
The contrastive learning based UPT~\cite{MoCo,chen2020simple,jia2021scaling,yang2022unified,oord2018representation,khosla2020supervised} trains the model by learning the prior knowledge distribution of the data itself by making similar instances closer and dissimilar instances farther apart in the feature space.
Recently, the autoencoder-based masked prediction methods have demonstrated great effectiveness on the large-scale foundation model pre-training. Such methods~\cite{MAE,xie2022simmim,li2022uniform,chen2022efficient,singh2022flava,BEiT,luo2022segclip} train the model to recover the masked image patches from a corrupted input image.
Our work is built on unsupervised unimodal pre-trained models, to fully leverage their generalization ability for boosting VLM tasks.

\noindent\textbf{Task Fine-tuning (TFT)}
is an essential step to transfer the high-capacity model from the pre-training domain to specific downstream tasks and applications by fine-tuning a subset of the model parameters on the target data, which is usually much less than pre-trained data.
Prompt tuning is a powerful TFT technique by optimizing learnable text prompts for Text Prompt Tuning~\cite{zhou2022learning,zhou2022conditional,ma2023understanding,derakhshani2022variational}, modulating the input of image encoder for Visual Prompt Tuning~\cite{jia2022visual,bahng2022visual}, and joint prompt optimization on multiple modalities for Text-Visual Prompt Tuning~\cite{zang2022unified,shen2022multitask,khattak2022maple,xing2022class}.
Feature Adaptation~\cite{gao2021clip,zhang2021tip,pantazis2022svl,houlsby2019parameter} is another popular TFT paradigm by fine-tuning additional light-weight feature adapters to adapt image or text features.
Some other TFT techniques explore cross attention~\cite{qiu2021vt,guo2022calip}, direct fine-tuning~\cite{wortsman2022robust} and architecture modification~\cite{zhou2021denseclip}.
Our work proposes a novel multitask fine-tuning method with a multiway multimodal neck to adapt the unsupervised pre-trained unimodal encoders to diverse downstream tasks.

\noindent\textbf{Multi-Task Learning (MTL)}
is a powerful learning paradigm to leverage the shared information of multiple related tasks~\cite{shen2022multitask,caruana1997multitask,dong2019unified,sun2019ernie,sun2020ernie,ren2018cross} to improve the generalization of all the tasks. MTL can optimize the model on multiple disjoint datasets, erasing the demanding for the expensive exhaustive labeling and simultaneously maintaining high performance on all tasks.
Existing approaches have been proposed for multitask learning, include hard parameter sharing, where the model shares all parameters across tasks, and soft parameter sharing, where the model shares some parameters across tasks while allowing other parameters to be task-specific.
Our UniBoost is a soft parameter sharing multitask framework whose input can be image, text or image-text pairs depending on the target task. 

\begin{figure}
    \centering
    \includegraphics[width=\linewidth]{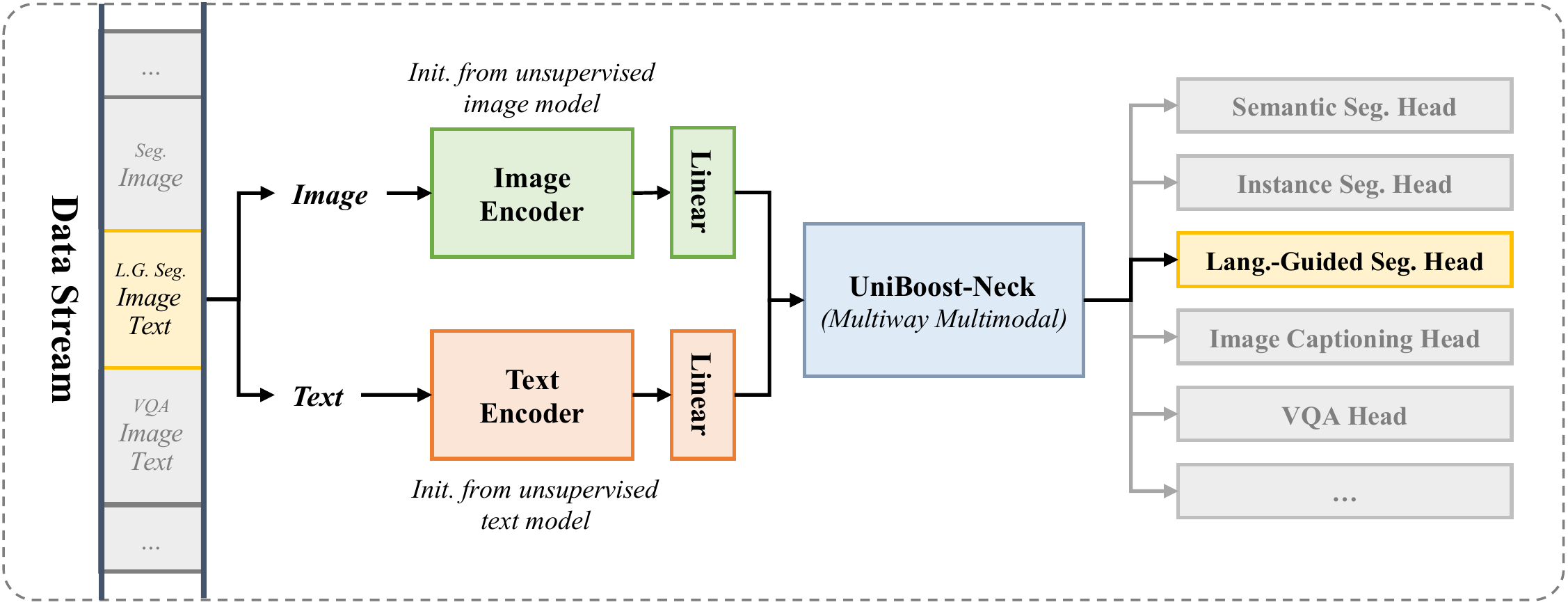}\vspace{-0.05in}
    \caption{{\bf UniBoost} is a {mul}titask {i}ntermediate {f}ine-tuning framework based on unsupervised unimodal encoders. The input can be image, text or image-text pair depending on the target task during multitask intermediate fine-tuning. The UniBoost-Neck is a multiway multimodal module as shown in Figure~\ref{fig:neck}}.
    \vspace{-0.2in}
    \label{fig:framework}
\end{figure}

\vspace{-0.1in}
\section{UniBoost: A Multitask Intermediate Fine-tuning Framework}
\vspace{-0.1in}
The pretrain-then-finetune paradigm is very effective which holds the holy grail for modern vision and vision-language tasks. 
Recently, intermediate fine-tuning has been inserted before fine-tuning the pre-trained models on the target task. Intermediate fine-tuning usually involves specific task objective and complex inference on a relatively large-scale dataset. Inspired by this, we propose \textbf{UniBoost}, a Multitask Intermediate Fine-tuning framework as shown in Figure~\ref{fig:framework} that can not only fully leverage the generalization ability of unsupervised unimodal models, but also make use of multitask in-domain data to optimize the alignment of image and text embedding space, thereby improving the ability of transferring knowledge to downstream tasks.

\begin{figure}
  \centering
  \includegraphics[width=1.0\textwidth]{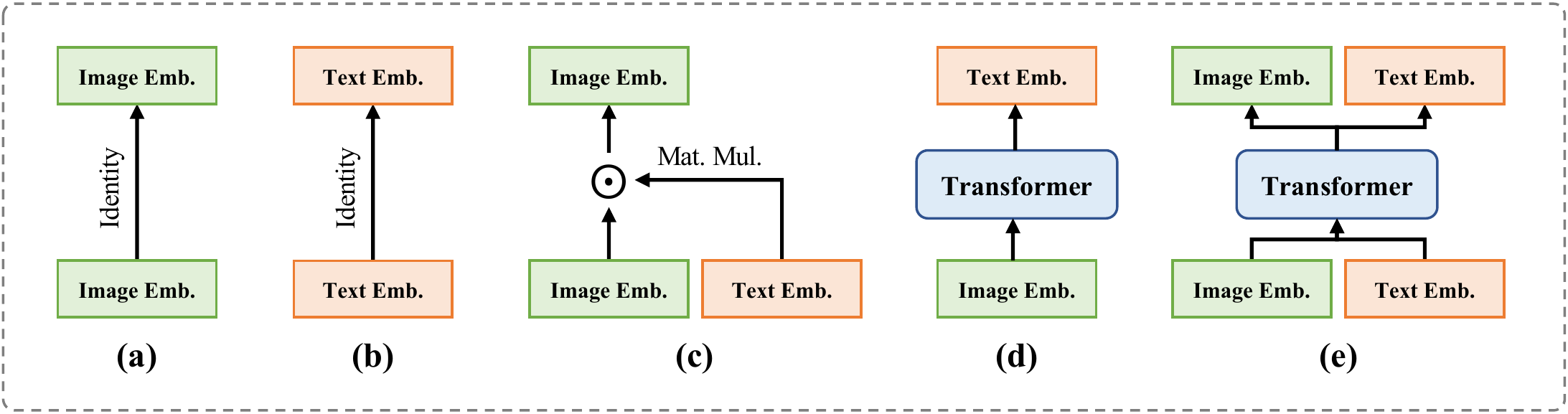}\vspace{-0.05in}
  \caption{\textbf{UniBoost-Neck} is a multiway multimodal module which supports five types of tasks: (a)--(b) Vision and language tasks which require image or text embedding only; (c) Language-guided vision tasks such as language-guided semantic segmentation; (d) Image-to-text generatation tasks such as image captioning; (e) Tasks which require deep fusion of vision and language embeddings such as VQA.} 
  \label{fig:neck} 
  \vspace{-0.2in}
\end{figure}

\subsection{Network}
We adopt separate image and text encoders with pre-trained weights from the unsupervised unimodal models for extracting image and text embeddings respectively. Next, a projection layer is employed on the extracted visual and textual embeddings separately to align the number of embedding dimension. Considering the framework will be trained on multiple tasks which require different levels of information, we take multi-layer features from the image encoder and concatenate them before sending to the projection layer. Then, we concatenate the projected image and text embeddings as a new sequence and feed it into the UniBoost-Neck for multimodal information interaction. The UniBoost-Neck is a multiway multimodal module which supports five types of tasks. 
As shown in Figure~\ref{fig:neck}, this neck supports five type of tasks: (a)--(b) unimodal tasks which require image or text embedding only, (c) language-guided vision tasks which require language-guided image embedding, (d) image-to-text generation tasks, or (e) multimodal tasks which require deep fusion of image and text embeddings. 
Afterward, the unimodal or multimodal representations are sent to task-specific heads to produce predictions and compute losses.

\subsection{Multitask Intermediate Fine-tuning}
Existing vision-language paradigms tend to learn representations for image and text, as well as their alignment (or fusion) jointly. As discussed in the introduction, such joint learning restricts the embedding space to the intersection of images and texts. Differently, we disjoint the representation and alignment (or fusion) learning. In UniBoost, since the image and text encoders are already initialized from well pre-trained unsupervised unimodal models, which adopts effective contrastive learning or masked modeling techniques on large-scale datasets, the extracted image and text embeddings are generalized and robust enough for transfer learning. Therefore, once the pre-trained unimodal encoders are settled, we just align (or fuse) the isolated image and text embedding space through further fine-tuning at a relatively low cost. Here, we directly fine-tune UniBoost on various in-domain tasks, namely \textbf{multitask intermediate fine-tuning}.  

\noindent{\textbf{Data.}} UniBoost is pre-trained on multiple in-domain tasks, involving high-level tasks such as classification, middle-level tasks such as semantic segmentation, low-level tasks such as denoising, vision-language tasks such as VQA, image captioning. In total, the in-domain pre-training is conducted on 12 tasks consisting of 6.5M images.

\noindent{\textbf{Multi-tasking.}} For the data stream, we arrange all tasks sequentially and group data points by batch for each task. Within each batch, the data points belong to the same task. For the task which cannot support sufficient samples in last batch, we resample from this dataset to make up an intact batch. In each round, we shuffle all the batches, which form the data queue popping data consumed by the model in each iteration. When the data queue is empty, we shuffle the data points within each task and repeat the aforementioned procedures to produce next batches. 

\noindent{\textbf{Experiment Setup.}} We train UniBoost for 1M iterations with a batch size of 64 on 8 cards of NVIDIA Tesla V100 with 32GB. UniBoost uses 16$\times$16 patch size and is trained at resolution 480$\times$480. We use the AdamW optimizer for optimization and a cosine learning rate schedule with a peak learning rate of 1e$^{-4}$ and a linear warmup of 5K steps. Note that the learning rate for encoders is multiplied by a ratio of 0.1. The weight decay is 0.01.

\vspace{-0.1in}
\section{Systematic Analysis}\label{sec:unimodal}
\vspace{-0.1in}
Our UniBoost provides more general and robust weights given by unsupervised unimodal pre-trained models when transferring knowledge to vision-language tasks. This section presents our systematic analysis and experiments to validate this argument on UniBoost with various  vision and language encoders across multiple tasks.

\subsection{Experiment Design}
In UniBoost, there are many choices of pre-trained models available, which can be grouped into three streams according to the supervision used in the pre-training stage, i.e., supervised pre-training, image-text pair supervised pre-training, unsupervised pre-training (or called self-supervised pre-training). 

Table~\ref{tab:encoders} shows that different pre-trained models adopt different pre-training data with different training strategies. For the image encoder, supervised pre-training models are trained on large-scale fundamental dataset with annotations, such as ViT~\cite{ViT} pre-trained on ImageNet-1K (IN1K). In image-text pair pre-training~\cite{CLIP}, models are trained through contrastive loss based on matched and unmatched image-text pairs in a large batch.  Various techniques are available for unsupervised pre-training. One of the most representative techniques is masked image modeling~\cite{MAE, BEiT} usually used in transformer-based model pre-training. For the text encoder, the situation is similar. Most popular pre-trained language models are trained through masked language modeling~\cite{T5} or contrastive learning from image-text pairs~\cite{CLIP}. Compared to supervised pre-training or image-text pair pre-training, unsupervised pre-training does not have any specific data requirements while is capable of learning more general and robust representations from a broader coverage of data distribution that can transfer well to new tasks and domains.

We transfer UniBoost with different pre-trained encoders to downstream vision-language tasks under zero-shot setting and supervised setting. To validate the effectiveness of unsupervised  unimodal pre-trained encoders, for each group of comparative experiments, we adopt the same architecture for UniBoost-Neck and task-specific head, same training hypermeters and schedule to ensure that only the image and text encoder are variables. In addition, in each group of experiments, we compare different pre-trained encoders with close number of model parameters to reduce the effect induced by model size. In this way, the performance gap will be caused mainly by the pre-trained weights rather than other factors. We conduct the comparison on several groups of different tasks and methods  detailed in the following.

\begin{table}[t]
\centering
\def\arraystretch{1.2}
\setlength{\tabcolsep}{2.0mm}{
\begin{tabular}{c|c|c|c|c|c|c}
\toprule
\multicolumn{2}{c|}{Model} & Data & Model Size & Layers & Width & Num Heads \\
\midrule
\multirow{9}{*}{\centering\arraybackslash Image Encoder} 
& ViT-B & IN1K & 86M & 12 & 768 & 12 \\
& CLIP-ViT-B & ITP & 86M & 12 & 768 & 12 \\
& MAE-B & IN1K & 86M & 12 & 768 & 12
\\
& BEiT-B & IN21K & 86M & 12 & 768 & 12 \\
\cmidrule{2-7}
& ViT-L & IN1K & 304M & 24 & 1024 & 16 \\
& CLIP-ViT-L & ITP & 304M & 24 & 1024 & 16 \\
& MAE-L & IN1K & 304M & 24 & 1024 & 16 \\
& BEiT-L & IN21K & 304M & 24 & 1024 & 16 \\
\midrule
\multirow{3}{*}{\centering\arraybackslash Text Encoder} 
& CLIP-B & ITP & 83M & 12 & 512 & 8 \\
& T5-S & C4 & 30M & 6 & 512 & 8 \\
& T5-B & C4 & 110M & 12 & 768 & 12 \\
\bottomrule
\end{tabular}}
\vspace{0.1in}
\caption{Pre-trained data, model size, and architecture of different pre-trained image and text encoders. ITP denotes image-text pair dataset used in CLIP~\cite{CLIP}.}
\label{tab:encoders}\vspace{-0.25in}
\end{table}

\subsection{Language-guided Semantic Segmentation}

\noindent{\textbf{Network.}} Language-guided semantic segmentation is a multimodal task which utilizes language information for pixel classification. Besides making use of an image encoder to generate image embedding, popular solution~\cite{lseg} of language-guided semantic segmentation task usually applies a text encoder to obtain embedding for the given classes, and then computes the similarity between the image embedding and the embedding of the given classes to produce predictions. Following this pipeline, we use the task-specific head designed in LSeg~\cite{lseg}, and DenseCLIP~\cite{DenseCLIP} equipped with UniBoost of different pre-trained image and text encoders, to conduct the comparison experiments. 

\noindent{\textbf{Dataset.}} We evaluate language-guided semantic segmentation equipped with UniBoost under two settings: zero-shot setting and supervised setting. We fine-tune and evaluate UniBoost with LSeg head under zero-shot setting on PASCAL-5$^i$ and COCO-20$^i$, which are popular few-shot segmentation datasets and are usually used in zero-shot segmentation methods to evaluate the generalization ability from seen classes to unseen classes. Specifically, PASCAL-5$^i$ split 20 classes into 4 folds, with each fold denoted as PASCAL-5$^i$, $i\in{1, 2, 3, 4}$. In each fold, 5 classes with the responding mask annotations are taken as the novel set in evaluation while the others form the base set used in training. Similarly, COCO-20$^i$ also have 4 folds of 20 classes each. For UniBoost with DenseCLIP head, we train and evaluate the model on ADE20K under supervised setting.

\noindent{\textbf{Experiment Setup.}} For UniBoost with LSeg head and different image and text encoders, we fine-tune each model for 15 epochs with a batch size of 8. Image augmentation includes random resized cropping, horizontal flipping, and color jittering. We use the SGD optimizer with a momentum of 0.9 for optimization following~\cite{lseg}. We use a linear learning rate schedule with a base learning rate of 5e$^{-5}$. For UniBoost with DenseCLIP head and different image and text encoders, we fine-tune each model for 80K steps with a batch size of 32. We use the AdamW optimizer for optimization, a poly learning rate schedule with a base learning rate of 1e$^{-4}$ and a linear warmup of 1500 steps. 

\begin{table}[t]
    \centering
    \def\arraystretch{1.1}
    \setlength{\tabcolsep}{2.8mm}{
    \begin{tabular}{c|ccc|ccc}
    \toprule
        \multicolumn{1}{c|}{\multirow{2}{*}{Split}} & 
        \multicolumn{3}{c|}{\multirow{1}{*}{Train}} & 
        \multicolumn{3}{c}{\multirow{1}{*}{Validation}} \\
        \cmidrule{2-7}
        \multicolumn{1}{c|}{\multirow{0}{*}{}} & 
        \multicolumn{1}{c}{\multirow{1}{*}{number}} & 
        \multicolumn{1}{c}{\multirow{1}{*}{yes/no}} & 
        \multicolumn{1}{c|}{\multirow{1}{*}{other}} & 
        \multicolumn{1}{c}{\multirow{1}{*}{number}} & 
        \multicolumn{1}{c}{\multirow{1}{*}{yes/no}} & 
        \multicolumn{1}{c}{\multirow{1}{*}{other}} \\
        \cmidrule{1-7}
        Base & 46243 & 131393& 175314& 22194 & 61693 & 82647 \\
        Novel & 11363 & 35489 & 43955 & 5940 & 18848 & 23032 \\
        \bottomrule
    \end{tabular}}
    \vspace{0.1in}
    \caption{Summary of zero-shot split on VQA v2.0 dataset.}
    \label{tab:VQAv2}\vspace{-0.25in}
\end{table}

\subsection{Language-guided Object Detection and Instance Segmentation}

\noindent{\textbf{Experiment Setup.}} Similar to language-guided semantic segmentation, language-guided object detection or instance segmentation methods utilize a text encoder to compute embeddings for the given classes, which are used as the classification weight for the detected objects. We adopt the instance segmentation head used in DenseCLIP~\cite{lseg} equipped with UniBoost of different pre-trained image and text encoders and fine-tune the model on COCO dataset under supervised setting. The maximum side of input image is set to 800 for model with transformer-based backbones while others adopt $1333\times800$. We train the model for a total of 12 epochs with a batch size of 16. We use the AdamW optimizer and a stepwise learning rate decay scheduler with a base learning rate of 2e$^{-4}$, which decays in epoch 8 and 11 with a ratio of 0.1.

\subsection{Visual Question Answering}

\noindent{\textbf{Network.}} 
For VQA, by considering it as an answer generation task rather than a multi-choice classification task, it allows for open-ended VQA. Specifically, we concatenate the embeddings of an image and a given question, and then feed them into the multimodal fusion module in the UniBoost-Neck, which is a 12-layer transformer with a width of 768 followed by a language model (LM) head for auto-regressive text generation.

\noindent{\textbf{Dataset.}} We evaluate VQA task under zero-shot setting by measuring open-ended accuracy. We build the dataset based on VQA v2.0 refer to the setting used in~\cite{fewshot_vqa_tao}.
VQA v2.0 consists of there answer types, i.e., ``number'', ``yes/no'' and ``other''. For the ``number'' and ``yes/no'' types, we gather the unique tokens present in questions, count their frequency and split them into base and novel sets. Specifically, tokens with frequency in [10, 40) are treated as novel while others belong to base set. For the ``other'' type, similar procedures are also applied, except slightly differently, we only take the tokens in all answers into consideration. A summary of the zero-shot dataset is tabulated in Table~\ref{tab:VQAv2}. Each model is trained on the base split of the train set and evaluated on the novel split of the validation set.

\noindent{\textbf{Experiment Setup.}}  
We train the model for a total of 10 epochs with a training batch size of 16. We use the AdamW optimizer, a cosine learning rate schedule with a peak learning rate of 2e$^{-5}$. The weight decay is 0.05. 

\vspace{-0.1in}
\section{Experimental Results}
\vspace{-0.05in}
In this section, we perform extensive evaluation on UniBoost with different task heads on public benchmarks for vision-language tasks. Notably, UniBoost with unsupervised pre-trained unimodal model outperforms recent state-of-the-art methods on a wide range of vision-language tasks by a large margin. We present the most significant results in the main paper, and defer additional comparisons to supplementary materials.

\begin{table*}[t]
    \centering
    \def\arraystretch{1.1}
    \setlength{\tabcolsep}{1.5mm}{
    \begin{tabular}{c|c|c|c|cccccc}
    \toprule
    Method & \makecell[c]{Image\\Encoder} & \makecell[c]{Text\\Encoder} & 
    Shot & 5$^0$ & 5$^1$ & 5$^2$ & 5$^3$ & mean & FB-IoU \\
    \midrule
    PFENet~\cite{PFENet} & \multirow{3}{*}{ResNet101} & - & 1-shot & 60.5 & 69.4 & 54.4 & 55.9 & 60.1 & 72.9 \\
    RePRI~\cite{RePRI} & & - & 1-shot & 59.6 & 68.6 & \textbf{62.2} & 47.2 & 59.4 & - \\
    HSNet~\cite{HSNet} & & - & 1-shot & \textbf{67.3} & \textbf{72.3} & 62.0 & \textbf{63.1} & \textbf{66.2} & \textbf{77.6} \\
    \midrule
    SPNet~\cite{SPNet} & ResNet101 & - & zero-shot & 23.8 & 17.0 & 14.1 & 18.3 & 18.3 & 44.3 \\
    ZS3Net~\cite{ZS3Net} & ResNet101 & - & zero-shot & 40.8 & 39.4 & 39.3 & 33.6 & 38.3 & 57.7 \\
    LSeg~\cite{lseg} & ViT-L & CLIP-B & zero-shot & 61.3 & 63.6 & 43.1 & 41.0 & 52.3 & 67.0 \\
    LSeg + UniBoost & MAE-L & CLIP-B & zero-shot & 67.3 & 65.1 & 46.7 & 47.3 & 56.6 & 69.4 \\
    LSeg + UniBoost & MAE-L & T5-S & zero-shot & \textbf{68.7} & \textbf{67.1} & \textbf{49.0} & \textbf{50.4} & \textbf{58.8} & \textbf{70.8} \\
    \toprule
    \end{tabular}}
    \vspace{-0.05in}
    \caption{Comparison of mIoU and FB-IoU for semantic segmentation on PASCAL-5$^i$. }
    \label{tab:exp_pascal}\vspace{-0.1in}
\end{table*}

\begin{table*}[t]
    \centering
    \def\arraystretch{1.1}
    \setlength{\tabcolsep}{1.5mm}{
    \begin{tabular}{c|c|c|c|cccccc}
    \toprule
    Method & \makecell[c]{Image\\Encoder} & \makecell[c]{Text\\Encoder} & 
    Shot & 20$^0$ & 20$^1$ & 20$^2$ & 20$^3$ & mean & FB-IoU \\
    \midrule
    PFENet~\cite{PFENet} & \multirow{2}{*}{ResNet101} & - & 1-shot & 36.8 & 41.8 & 38.7 & 36.7 & 38.5 & 63.0 \\
    HSNet~\cite{HSNet} & & - & 1-shot & \textbf{37.2} & \textbf{44.1} & \textbf{42.4} & \textbf{41.3} & \textbf{41.2} & \textbf{69.1} \\
    \midrule
    ZS3Net~\cite{zsseg} & ResNet101 & - & zero-shot & 18.8 & 20.1 & 24.8 & 20.5 & 21.1 & 55.1 \\
    LSeg & ViT-L & CLIP-B & zero-shot & 28.1 & 27.5 & 30.0 & 23.2 & 27.2 & 59.9 \\
    LSeg + UniBoost & MAE-L & CLIP-B & zero-shot & 30.4 & 31.8 & 35.7 & 33.5 & 32.9 & 61.9 \\
    LSeg + UniBoost & MAE-L & T5-S & zero-shot & \textbf{31.0} & \textbf{33.2} & \textbf{35.9} & \textbf{33.6} &  \textbf{33.4} & \textbf{62.3} \\
    \toprule
    \end{tabular}}
    \vspace{-0.05in}
    \caption{Comparison of mIoU and FB-IoU for semantic segmentation on COCO-20$^i$.}
    \label{tab:exp_coco}\vspace{-0.2in}
\end{table*}

\noindent{\textbf{Language-guided Semantic Segmentation.}}
We test our method for language-guided semantic segmentation on two methods, LSeg \cite{lseg} for zero-shot semantic segmentation, and DenseCLIP \cite{DenseCLIP} for fully-supervised close-set semantic segmentation.

In Table~\ref{tab:exp_pascal}, we show the results of UniBoost on LSeg by replacing its vision and language encoders with our UniBoost, and by utilizing different pre-trained image and text encoders to validate the substantial improvement of unsupervised pre-training over supervised pre-training or image-text pair supervised pre-training. We find that the models with unsupervised pre-trained weights consistently outperform the model with supervised pre-trained weights or image-text pair supervised pre-trained weights.
Specifically, based on the supervised pre-trained model, if we replace the image encoder of ViT-L~\cite{ViT} pre-trained on ImageNet-1K with an unsupervised pre-trained image encoder, i.e., MAE-L~\cite{MAE}, the performance is improved to 56.6\%. Then, if we further utilize T5~\cite{T5} as the pre-trained text encoder, the performance is boosted to 58.8$\%$. Note that our zero-shot results even out-perform the 1-shot results by HSNet~\cite{HSNet} in 5$^0$.
Table~\ref{tab:exp_coco} shows the evaluation results on COCO-20$^i$ dataset. With our UniBoost, the performance is boosted from 27.2\% to 33.4\%. These experiments demonstrate the effectiveness of our proposal on aligning unsupervised pre-trained unimodalities for better performance on zero-shot tasks.

Similar conclusion can be drawn for fully-supervised experiments on DenseCLIP \cite{DenseCLIP}, as tabulated in Table \ref{tab:exp_ade20k}, where we replace the CLIP-based pre-trained weights by the unsupervised pre-trained weights on more data for both image encoder and text encoder, improving the performance DenseCLIP from 50.6\% to 52.9\% mIoU on ADE20K under fully supervised setting. This also demonstrates the effectiveness of unsupervised pre-training compared to image-text pair supervised pre-training.

\begin{table*}[tp]
    \centering
    \def\arraystretch{1.1}
    \setlength{\tabcolsep}{2.0mm}{
    \begin{tabular}{c|c|c|cc}
    \toprule
    Method & Image Encoder & Text Encoder & mIoU & pixAcc \\
    \cmidrule{1-5}
    SETR~\cite{SETR} & ViT-B & - & 46.2 & - \\ 
    Semantic FPN~\cite{PanopticFPN} & ViT-B & - & 49.1 & -\\
    BEiT3~\cite{beit3} + FPN & BEiT3-B & - & 48.4 & 83.2 \\
    DenseCLIP~\cite{DenseCLIP} & CLIP-ViT-B & CLIP-B & 50.6 & -\\
    DenseCLIP + UniBoost & BEiT-B & T5-B & \textbf{52.9} & \textbf{84.6} \\
    \bottomrule
    \end{tabular}}
    \caption{Comparison of mIoU and pixAcc for semantic segmentation on ADE20K under fully supervised setting. All models are evaluated under single scale.}
    \label{tab:exp_ade20k}\vspace{-0.15in}
\end{table*}

\begin{table*}[tp]
    \centering
    \def\arraystretch{1.1}
    \setlength{\tabcolsep}{1.2mm}{
    \begin{tabular}{c|c|c|cccccc}
    \toprule
     Method & Image Encoder & Text Encoder & 
    AP$^b$ & AP$^b_{50}$ & AP$^b_{75}$ & AP$^m$ & AP$^m_{50}$ & AP$^m_{75}$\\
    \midrule
    DenseCLIP~\cite{DenseCLIP} & CLIP-RN101 & CLIP-B & 42.6 & 65.1 & 46.5 & 39.6 & 62.4 & 42.4 \\
    DenseCLIP~\cite{DenseCLIP} & CLIP-ViT-B & CLIP-B & 41.3 & 64.1 & 44.5 & 37.8 & 60.7 & 39.8 \\
    DenseCLIP + UniBoost & BEiT-B & T5-B & \textbf{44.3} & \textbf{67.3} & \textbf{48.8} & \textbf{40.8} & \textbf{63.8} & \textbf{43.3} \\
    \bottomrule
    \end{tabular}}
    \caption{Comparisons of box AP and mask AP for object detection and instance segmentation respectively on COCO under fully supervised setting. DenseCLIP with CLIP-RN101 is trained and evaluated with the input size of $1333\times800$ while the other transformer-based models use the input size with a maximum side of 800 due to the memory limitation.}
    \label{tab:exp_coco2}\vspace{-0.15in}
\end{table*}

\begin{table*}[tp]
    \centering
    \def\arraystretch{1.1}
    \setlength{\tabcolsep}{1.8mm}{
    \begin{tabular}{c|c|c|c|c}
    \toprule
     Method & Setting & Image Encoder & Text Encoder & 
     mAP \\
    \midrule
    Faster R-CNN~\cite{FasterRCNN} & Inductive  & VGG & - & 73.2 \\
    Faster R-CNN~\cite{FasterRCNN} & Inductive  & ResNet101 & - & 75.2 \\
    YOLO v2~\cite{YOLO9000} & Inductive & Darknet-19 & - & 78.6 \\
    CenterNet~\cite{CenterNet} & Inductive & DLA-34 & - & \textbf{80.7} \\
    \midrule
    DenseCLIP~\cite{DenseCLIP} & Transductive & CLIP-ViT-B & CLIP-B & 74.1 \\
    DenseCLIP + UniBoost & Transductive & BEiT-B & T5-B & \textbf{77.4} \\
    \bottomrule
    \end{tabular}}
    \caption{Comparisons of mAP for object detection on PASCAL VOC 2007. Models under transductive setting are trained on COCO dataset.}
    \label{tab:exp_transfer}\vspace{-0.1in}
\end{table*}

\begin{table*}[t]
    \centering
    \def\arraystretch{1.1}
    \setlength{\tabcolsep}{2.2mm}{
    \begin{tabular}{c|c|c|cccc}
    \toprule
    Method & Image Encoder & Text Encoder & 
    number & yes/no & other & mean \\
    \midrule
    Baseline & CLIP-ViT-B & CLIP-B & 30.3 & 73.5 & 22.3 & 42.1 \\
    UniBoost + LM Head & BEiT-B & T5-B & \textbf{34.1} & \textbf{75.9} & \textbf{26.0} & \textbf{45.3} \\
    \bottomrule
    \end{tabular}}
    \caption{Comparisons of open-ended accuracy for visual question answering on VQA v2.0 dataset.}
    \label{tab:vqa_exp}\vspace{-0.2in}
\end{table*}

\noindent{\textbf{Language-guided Object Detection and Instance Segmentation.}}
Compared to semantic segmentation, instance segmentation can distinguish  different instances of the same semantic category. We  test our model for language-guided object detection and instance segmentation to further demonstrate its effectiveness. Table~\ref{tab:exp_coco2}  evaluates UniBoost with DenseCLIP head on COCO dataset. Compared to the model with image-text pair pre-trained weights, our UniBoost with unsupervised pre-trained unimodal BEiT-B and T5 outperforms by 1.7\% (42.6\%$\rightarrow$44.3\%) and 1.2\% (39.6\%$\rightarrow$40.8\%) box AP and mask AP under supervised setting, respectively, especially ours adopts a smaller input size. 

Additionally, we conduct transductive experiments to evaluate the generalization performance of our model on object detection. In our transductive experiment, the models are trained on the COCO dataset, which contains a large number of object detection images with diverse object categories and backgrounds. We then evaluate the model on the  the PASCAL VOC 2007, which contains a different set of object categories and backgrounds compared to COCO. By evaluating the models on a different dataset than the one used for training, we can gain insights into how well the models can generalize to new, unseen examples. This is particularly important in object detection, where the ability to detect objects accurately in a wide range of scenarios is critical for real-world applications. As shown in Table~\ref{tab:exp_transfer}, our UniBoost achieves larger performance promotion by 3.3\% (74.1\%$\rightarrow$77.4\%) compared to the CLIP-based model and are even comparable with some supervised models. 

\noindent{\textbf{Visual Questioning Answering.}}
VQA, as the most typical vision-language task, requiring the model to simultaneously understand the context of images and texts, and thus can be used to evaluate the capability of a vision-language model. In Table~\ref{tab:vqa_exp}, we validate the effectiveness of UniBoost compared to the model with the same architecture but initialized from CLIP model named as ``baseline''. Our result shows that, on the task of VQA,  unsupervised pre-trained unimodal models still outperforms image-text pair pre-trained models. Specifically, with the image encoder pre-trained by BEiT-B on ImageNet-21K dataset and the text encoder of T5 pre-trained on C4 dataset, our UniBoost achieves 3.2\% (42.1\%$\rightarrow$45.3\%) higher average accuracy on novel split of the VQA v2.0 validation set. 

All of the above improvements validate our claim that in transfer learning to zero-shot or supervised vision-language tasks, aligning two isolated spaces is easier and more effective than adapting a well-trained image-text alignment space, shown by better performance achieved by our new model on downstream vision-language tasks.

\vspace{-0.1in}
\section{Limitation} 
\vspace{-0.05in}
While our current framework and experiments focus on image and text data, it is important to note that the latest multimodal models, such as ImageBind~\cite{ImageBind}, include audio and other modalities in their embedding space. By incorporating other modalities, these models are able to capture a wider range of contextual information, leading to more different types of applications. Although our proposed framework has shown promising results in enhancing the zero-shot performance of vision-language tasks, it is important to continue studying the applicability of our method to other modalities. We believe that the inclusion of other modalities can further improve the quality and diversity of pre-training. This can potentially lead to even better performance in downstream tasks by enabling the model to better understand the complex relationships between different modalities via our UniBoost framework.

\vspace{-0.1in}
\section{Conclusion} 
\vspace{-0.05in}
In this paper, we validate that unsupervised unimodal pre-training can significantly boost the performance of zero-shot vision-language tasks, in comparison against supervised pre-training or image-text pair pre-training. In principle,  unsupervised pre-training can make use of not only image-text pair data or their intersection, but also image and text data on their own, the union of which cover a much broader and more diverse data distribution compared to image-text pairs. To take this advantage, we introduce UniBoost, a multitask intermediate fine-tuning framework based on unsupervised unimodal encoders. UniBoost leverages the generalization power of unsupervised unimodal embedding and can learn a broader joint image-text solution space by incorporating multi-task in-domain data, thereby improving zero-shot performance on downstream tasks. 
Extentive experiments demonstrate that equipped with our UniBoost based on  unsupervised unimodal pre-trained models, many methods such as LSeg and DenseCLIP achieve new state-of-the-art performance under zero-shot setting as well as supervised setting.

\vspace{-0.1in}
\section{Broader Impacts} 
\vspace{-0.05in}
UniBoost, unsupervised unimodal pre-training for
boosting zero-shot learning, relaxes the requirement of aligning  different modalities (e.g. text-image pair-up in CLIP). This will substantially benefit future downstream multimodal tasks, where individual unsupervised pretrained unimodal models can be readily deployed, which are respectively trained on broad data distributions. 

\newpage
\begin{appendices}
\vspace{-0.1in}
\section{Implementation Details}
\begin{figure}[t]
    \centering
    \includegraphics[width=1.0\linewidth]{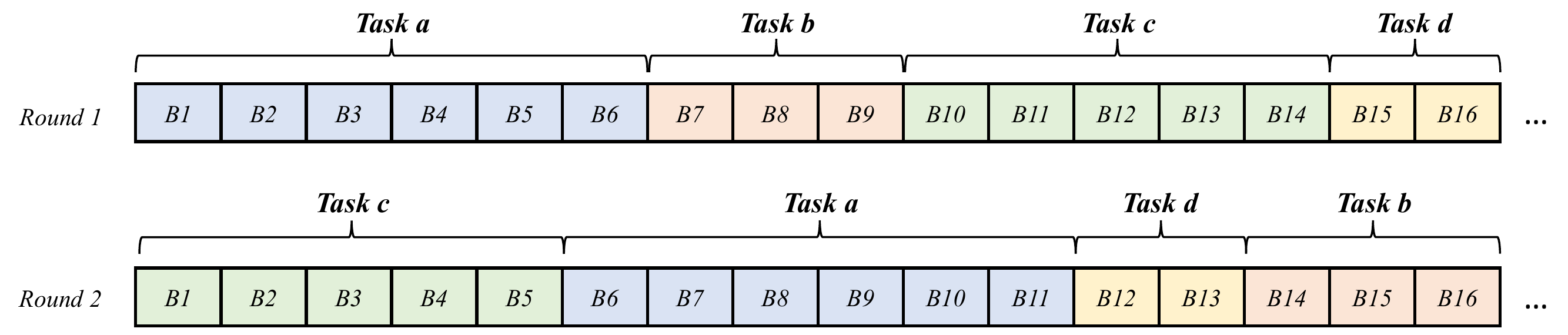}
    \caption{Data stream of multitasking.}
    \label{fig:datastream}
\end{figure}

\vspace{-0.1in}
\noindent{\textbf{Data and multi-tasking.}} Our UniBoost is fine-tuned on multiple tasks as summarized  in Table~\ref{tab:tasks} to learn the alignment or fusion between visual and textural embedding space. During training, we group all the data by batch and iterate batches in each round as shown in Figure~\ref{fig:datastream}. 
To balance the weights of different tasks during training, especially for those with a small number of total training images, e.g., less than 64,000 images, we  augment the images corresponding to these tasks multiple times by random crop and random resizing with scales ranging from 0.8 to 1.2, so that the number of total training images is around 64,000 at least which is an empirical value.

\begin{table}[htp]
    \def\arraystretch{1.1}
    \centering
    \setlength{\tabcolsep}{5.0mm}{
    \begin{tabular}{l|c|c}
        \toprule
        Task & Dataset & Number of Images \\
        \midrule
        Image classification & ImageNet-1K~\cite{ImageNet} & 1,000,000 \\
        \midrule
        \multirow{2}{*}{Object detection} & COCO~\cite{coco} & 118,287 \\
         & Objec365~\cite{objects365} & 2,000,000 \\
        \midrule
        Instance segmentation & COCO~\cite{coco} & 118,287 \\
        \midrule
        \multirow{2}{*}{Semantic segmentation} &  ADE20K~\cite{ade} & 20,000 \\
        & PASCAL Context~\cite{pascal_context} & 10,103 \\
        \midrule
        Language-guided detection & COCO~\cite{coco} & 118,287 \\
        \midrule
        \multirow{3}{*}{Language-guided segmentation} & ADE20K~\cite{ade} & 20,000 \\
        & PASCAL Context~\cite{pascal_context} & 10,103 \\
        & FSS1000~\cite{FSS1000} & 10,000 \\
        \midrule
        Depth estimation & NYUv2~\cite{NYUv2} & 1,449 \\
        \midrule
        Denoising & SIDD~\cite{SIDD} & 30,000 \\
        \midrule
        Deblurring & GoPro~\cite{GoPro} & 3214 \\
        \midrule
        \multirow{4}{*}{Image captioning} & COCO~\cite{coco} & 82,783 \\
         & Nocaps~\cite{nocaps} & 15,100 \\
         & SBU captions~\cite{SBU} & 860,000 \\
         & Conceptual Captions~\cite{CC} & 1,000,000 \\
        \midrule
        \multirow{2}{*}{Visual questioning answering} & VQA v2.0~\cite{VQAv2} & 82,783 \\
         & Visual Genome~\cite{VG} & 101,174 \\
        \midrule
        Visual Reasoning & NLVR2~\cite{NLVR2} & 107,292 \\
        \bottomrule
    \end{tabular}
    }
    \vspace{0.1in}
    \caption{Tasks and datasets used for multitask intermediate fine-tuning.}
    \label{tab:tasks}
\end{table}

\vspace{-0.15in}
\section{More Experimental Results}
\begin{figure}[htp]
    \centering
    \includegraphics[width=1.0\linewidth]{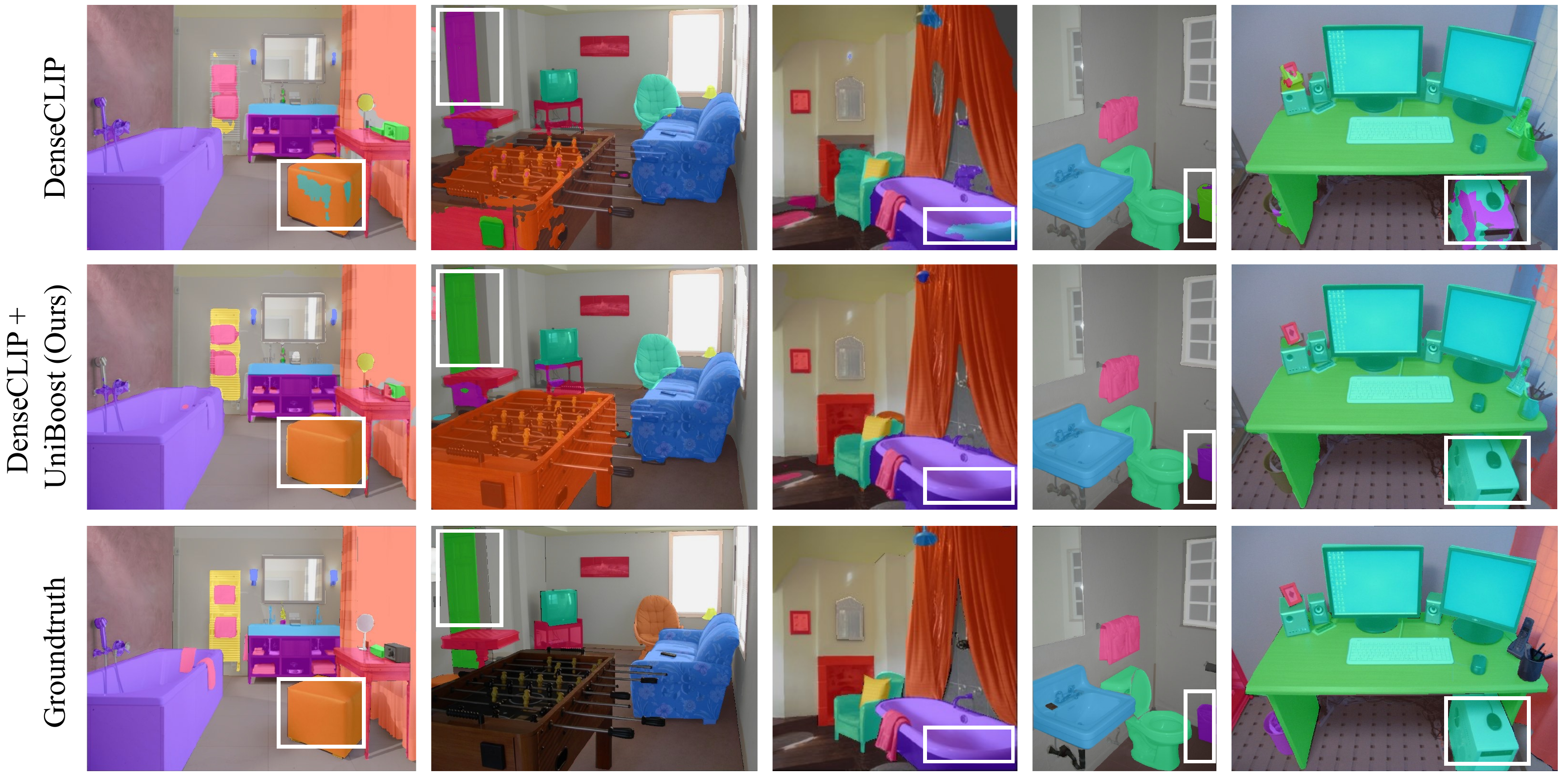}
    \caption{Qualitative results on ADE20K.}
    \label{fig:vis1}
\end{figure}

\begin{figure}
    \centering
    \includegraphics[width=1.0\linewidth]{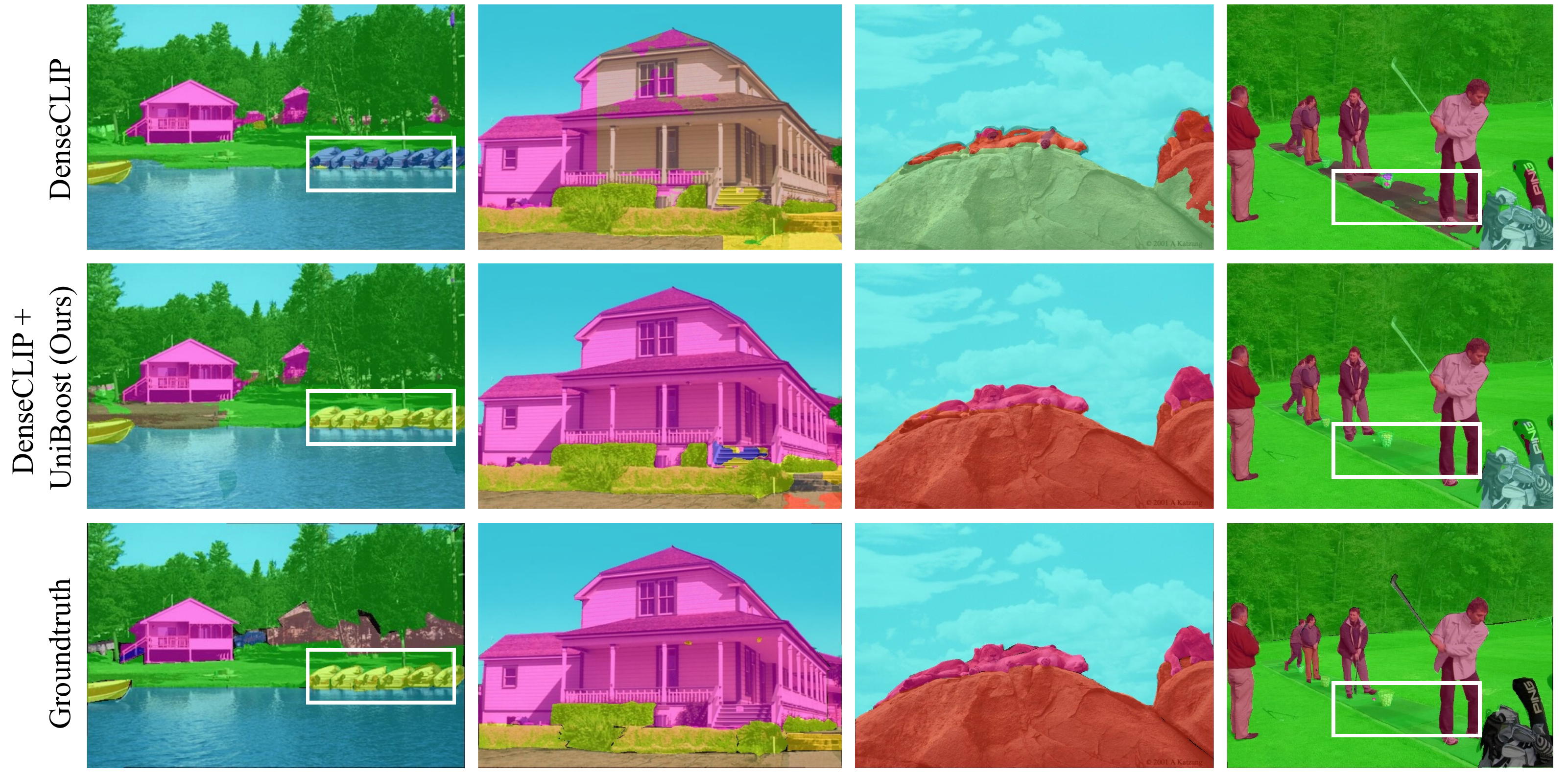}
    \caption{Qualitative results on ADE20K.}
    \label{fig:vis2}
\end{figure}

\begin{figure}
    \centering
    \includegraphics[width=1.0\linewidth]{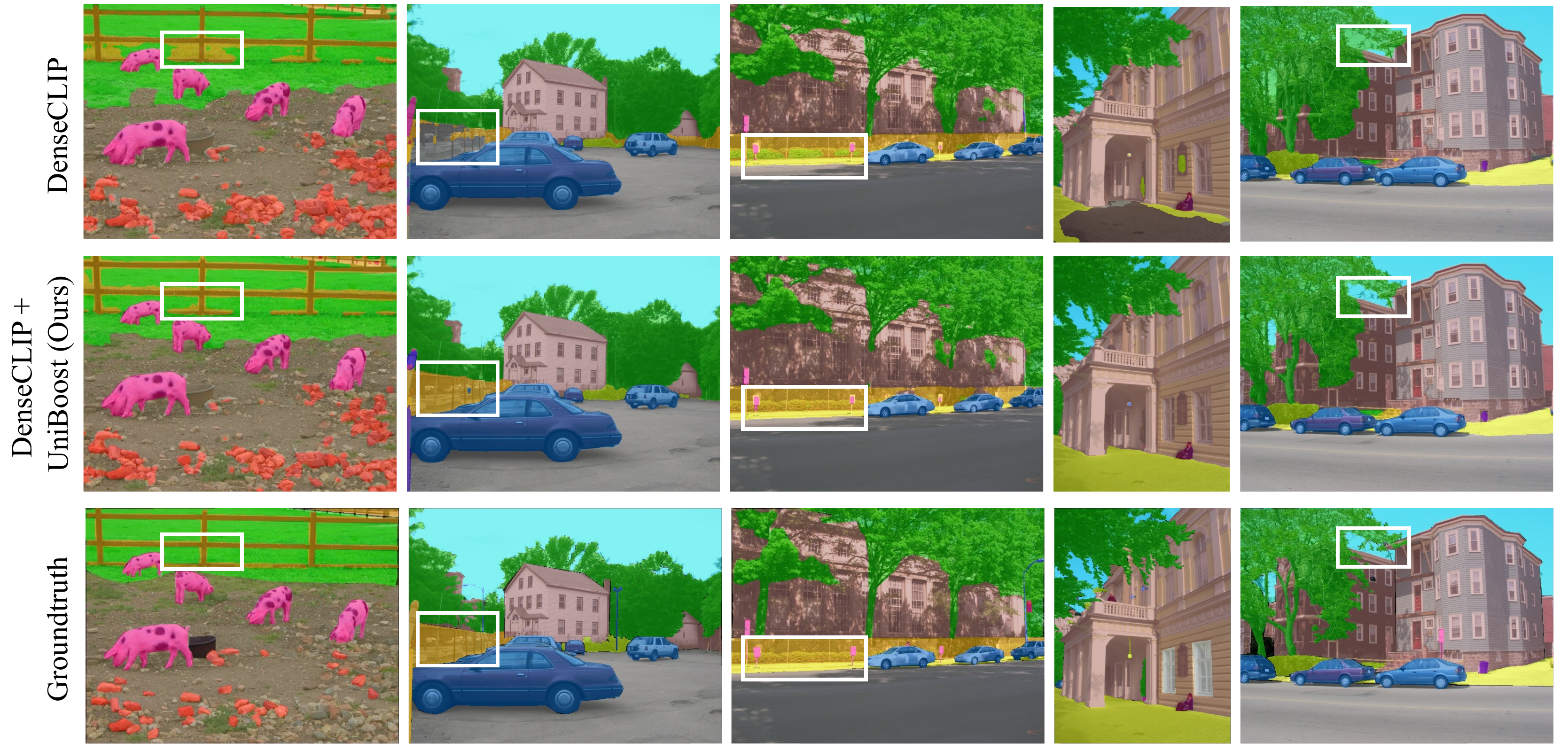}
    \caption{Qualitative results on ADE20K.}
    \label{fig:vis3}
\end{figure}

\vspace{-0.1in}
\noindent{\textbf{Language-Guided Semantic Segmentation.}} We give the zero-shot results of language-guided semantic segmentation evaluated on PASCAL-5$^i$ and COCO-20$^i$.
To test the robustness and generalization ability of zero-shot semantic segmentation models across different domains of datasets, we evaluate the models trained on different split of PASCAL-5$^i$ on COCO dataset. Since PASCAL VOC and COCO data share some classes, we remove all the common classes during evaluation to make sure all the evaluated classes are novel to the model. Table~\ref{tab:pascal_coco} tabulates the results, where we provide the performance of LSeg with CLIP-RN101 for reference since the weights of LSeg with ViT-L are not available. As shown, our models trained on different splits of PASCAL-5$^i$ achieves higher and more stable performance on COCO dataset as LSeg exhibits a large gap among the models trained on different splits. Additionally, we provide qualitative results in multiple scenarios including indoor, landscape and street views on ADE20K dataset in Figure~\ref{fig:vis1}--\ref{fig:vis3}. As shown, our method boosts the classification or pixel grouping accuracy, thus improves the segmentation performance benefiting from more general and robust representation.

\noindent{\textbf{Visual Questioning Answering.}} Besides the comparisons under zero-shot setting, we also present the results under supervised setting on VQA v2.0 dataset. Under supervised setting, we treat VQA as a multi-choice classification task. Specifically, we collect all unique answers as the candidates and force the model to select the correct answer for a given question, which is implemented by a classification head. Table~\ref{tab:VQAv2_test} summarizes the results.

\noindent{\textbf{Image Captioning.}}
Image caption is a task of generating natural language description for an image. During training, we model this task as a conditional generation task with masked signal modeling, which predicts the masked tokens through unmasked context. To be specific, we randomly drop partial ground-truth caption words, and train the model to recover these masked tokens based on the clues of images and leftward caption context, which is supervised by the cross entropy loss. For attention operation, text tokens are only capable of attending to image tokens, their leftward text tokens, and themselves. During inference, we generate the caption tokens in auto-regressive manner. We fine-tune an evaluate our model on Nocaps dataset, and report CIDEr on the validation set presented in Table~\ref{tab:nocaps}. 

\begin{table}[t]
    \def\arraystretch{1.2}
    \centering
    \setlength{\tabcolsep}{2.2mm}{
    \begin{tabular}{c|c|c|ccccc}
    \toprule
         \multirow{2}{*}{Method} & 
         \multirow{2}{*}{\makecell[c]{Image\\Encoder}} & 
         \multirow{2}{*}{\makecell[c]{Text\\Encoder}} & 
         \multicolumn{5}{c}{PASCAL-5$^i$ $\rightarrow$ COCO}\\
         & & & 5$^0$ & 5$^1$ & 5$^2$ & 5$^3$ & mean \\
    \midrule
         LSeg & CLIP-RN101 & CLIP-B & 24.93 & 23.88 & 17.03 & 27.04  & 23.22 \\
         LSeg + UniBoost & MAE-B & T5-S & \textbf{26.66} & \textbf{28.04} & \textbf{27.10} & \textbf{28.60} & \textbf{27.60} \\
    \bottomrule
    \end{tabular}}
    \vspace{0.1in}
    \caption{Evaluation on COCO of language-guided semantic segmentation models trained on different splits of PASCAL-5$^i$. Since the weights of LSeg with CLIP-ViT-L are not available, we provide the performance of LSeg with CLIP-RN101 for reference.}
    \label{tab:pascal_coco}
    \vspace{-0.1in}
\end{table}

\begin{table}[htp]
    \def\arraystretch{1.2}
    \centering
    \setlength{\tabcolsep}{2.7mm}{
    \begin{tabular}{c|c|c|cc}
    \toprule
         Method & \makecell[c]{Image Encoder} & \makecell[c]{Text Encoder} & test-dev & test-std\\
    \midrule
         baseline & CLIP-ViT-B & CLIP-B & 75.51 & 75.64 \\
         UniBoost + LM head & BEiT-B & T5-B & 77.75 & 77.92 \\
    \bottomrule
    \end{tabular}}
    \vspace{0.1in}
    \caption{Visual questioning answering on VQA v2.0 test set.}
    \label{tab:VQAv2_test}

    \def\arraystretch{1.2}
    \centering
    \setlength{\tabcolsep}{2.7mm}{
    \begin{tabular}{c|c|c|c}
    \toprule
         Method & \makecell[c]{Image Encoder} & \makecell[c]{Text Encoder} & CIDEr\\
    \midrule
         baseline & CLIP-ViT-B & CLIP-B & 101.7 \\
         UniBoost + LM head & BEiT-B & T5-B & 103.1 \\
    \bottomrule
    \end{tabular}}
    \vspace{0.1in}
    \caption{Image captioning on Nocaps validation set.}
    \label{tab:nocaps}
\end{table}

\end{appendices}

\newpage
{\small
\bibliographystyle{unsrtnat}
\bibliography{egbib}
}


\end{document}